\documentclass[conference]{IEEEtran}
\IEEEoverridecommandlockouts

\usepackage{cite}
\usepackage{amsmath,amssymb,amsfonts}
\usepackage{algorithmic}
\usepackage{graphicx}
\usepackage{textcomp}
\usepackage{xcolor}
\usepackage{tikz}
\usepackage{enumitem}
\usepackage{booktabs}
\usepackage{hyperref}
 
\usepackage{float}
\usepackage[ruled,vlined]{algorithm2e}

\def\BibTeX{{\rm B\kern-.05em{\sc i\kern-.025em b}\kern-.08em
    T\kern-.1667em\lower.7ex\hbox{E}\kern-.125emX}}
\newcommand*\circled[1]{\tikz[baseline=(char.base)]{%
  \node[shape=circle, fill=blue!20, draw, inner sep=2pt] (char) {#1};}}

\usepackage{wasysym}
\usepackage{etoolbox}

\AtBeginEnvironment{itemize}{
    
    }

\begin{document}

\title{Fusion-ResNet: A Lightweight multi-label NILM Model Using PCA–ICA Feature Fusion

{\footnotesize \textsuperscript{}}
\thanks{}
}

\author{
\IEEEauthorblockN{Sahar Moghimian Hoosh$^{1,2,a}$, Ilia Kamyshev$^{1,2,b}$, Henni Ouerdane $^{1,c}$}
\IEEEauthorblockA{$^{1}$ Center for Digital Engineering, Skolkovo Institute of Science and Technology, Moscow, Russian Federation \\
$^{2}$ Monisensa Development LLC, Moscow, Russian Federation\\
$^{a}$sahar.moghimian@skoltech.ru, $^{b}$ilia.kamyshev@skoltech.ru} $^{c}${h.ouerdane@skoltech.ru} }

\maketitle
\thispagestyle{plain}
\pagenumbering{arabic}

\begin{abstract}

Non-intrusive load monitoring (NILM) is an advanced load monitoring technique that uses data-driven algorithms to disaggregate the total power consumption of a household into the consumption of individual appliances. However, real-world NILM deployment still faces major challenges, including overfitting, low model generalization, and disaggregating a large number of appliances operating at the same time. To address these challenges, this work proposes an end-to-end framework for the NILM classification task, which consists of high-frequency labeled data, a feature extraction method, and a lightweight neural network. Within this framework, we introduce a novel feature extraction method that fuses Independent Component Analysis (ICA) and Principal Component Analysis (PCA) features. Moreover, we propose a lightweight architecture for multi-label NILM classification (Fusion-ResNet). The proposed feature-based model achieves a higher $F1$ score on average and across different appliances compared to state-of-the-art NILM classifiers while minimizing the training and inference time. Finally, we assessed the performance of our model against baselines with a varying number of simultaneously active devices. Results demonstrate that Fusion-ResNet is relatively robust to stress conditions with up to 15 concurrently active appliances. 
\end{abstract}

\vspace{10pt} 

\begin{IEEEkeywords}
Non-intrusive load monitoring (NILM), multi-label classification, energy disaggregation, appliance recognition, independent component analysis (ICA), principal component analysis (PCA)
\end{IEEEkeywords}

\section{Introduction}
\label{submission}
Non-intrusive load monitoring, also known as energy disaggregation, is the technique of breaking down a household's total energy use into individual appliance-level components using advanced analysis\cite{angelis2022nilm}. This technique has received significant attention in recent years due to its potential to enable greater energy efficiency, demand response, and load forecasting \cite{Event_detection}. The concept of NILM was first introduced in the 1980s by G. Hart \cite{hart1992nonintrusive}, and since then, it has been a popular topic of research in the field of electrical energy management. 

Various techniques have been suggested to enhance the accuracy of NILM, which, however, can be affected by several factors such as the total number of appliances, the number of appliances working simultaneously, the types of appliances, and the measurement noise. Multiple simultaneous appliance switching detection and correct estimation in practical scenarios with noisy data, remain to be addressed. Recent studies of deep learning models for NILM \cite{en13092195} indicate that training deep neural networks on limited labeled data can reduce the disaggregation accuracy, and model generalization  or lead to overfitting \cite{angelis2022nilm}. Using an excessive number of features can also cause overfitting. Typical algorithms are heavy in memory and sensitive to overfitting, which makes them difficult to be ported to sensors \cite{chang2015feature}. Moreover, most of the algorithms are trained on a limited number of appliances, while datasets contain dozens of classes of appliances on average. There is also a lack of related studies on the ``goodness’’ of disaggregation, with numerous individual components presented in the aggregated signal. The available datasets have limited combinations of different appliances, biased towards the most frequently used ones \cite{kamyshev2025cold}.  Thus, algorithmic complexity, limited labeled datasets, and the challenge of selecting a rich yet compact feature set are considered as major obstacles in the field of energy disaggregation \cite{Rafiq2021GeneralizabilityIO}. These challenges become critical when dealing with high-frequency NILM datasets, which are rare and often come with certain limitations. All these issues must to be addressed to improve the accuracy and efficiency of energy disaggregation algorithms. 

In our previous work \cite{Sahar}, to minimize NILM computational load, we developed a neural network architecture that uses independent component analysis (ICA) to extract features from high-frequency data. Here, we propose an extension of this work as a lightweight multi-label classifier with a novel feature  extraction method that fuses independent component analysis and principal component analysis (PCA) features. We compare the proposed algorithm and state-of-the-art NILM classification models for a varying number of individual appliances operating simultaneously. We demonstrate that the proposed model is less prone to overfitting, exhibits low complexity, and performs well when there are numerous appliances working at the same time (up to 15).  To the best of our knowledge, this research is the first to use the fusion of ICA and PCA features to enhance the performance of the multi-label classification models in NILM, specifically with high-frequency sampling data ($>1$ kHz). The results of this study provide valuable insights for the development of accurate and efficient energy disaggregation algorithms capable of handling complex scenarios and diverse datasets. This work is implemented in Python and is available via the provided \href{https://github.com/arx7ti/ML2023SK-final-project}{link}. 

The article is organized as follows: In Section \ref{sec:related_works}, we review related works, focusing on existing feature extraction methods and their limitations. In Section \ref{sec:dataset}, we describe the dataset and the preprocessing procedure in detail. The fusion strategy for feature extraction and mathematical formulation of PCA and ICA are presented in Section \ref{sec:features}. In Section \ref{sec:algorithms}, we present our proposed multi-label classification model (Fusion-ResNet) and baseline algorithms. The experimental setup and the evaluation of models under different numbers of simultaneously operating appliance, are discussed in Sections \ref{sec:experiments} and \ref{sec:results}. Finally, In Section \ref{sec:conclusion}, we summarize our main findings and outline future research directions. 

\section{Related Works}\label{sec:related_works}
The first step of NILM algorithms development is to select or extract the most informative features from the original datasets. In theory, all available features can be used as inputs to NILM algorithms; however, irrelevant or highly correlated features may cause overfitting and reduce the model’s ability to generalize to unseen data. The study of feature extraction for NILM has evolved from simple power-based indicators to sophisticated high-frequency signal representations capturing temporal, spectral, and geometric information. Early research works such as \cite{sultanem2002using} predominantly used steady-state power quantities—active, reactive, and apparent power—because they were easily obtainable from voltage and current measurements and computationally affordable to calculate. However, as Hart demonstrated \cite{hart1992nonintrusive}, these quantities are not sufficient to distinguish between appliances with similar power consumption. Hart therefore proposed extending the feature space to include harmonic content and transient behavior. Later, Leeb et al. \cite{leeb1995transient} introduced wavelet and spectral-envelope analyses to localize transient events in both time and frequency. Their motivation was to capture the device switching events. Voltage–current (V–I) trajectory features were proposed by Ting et al. \cite{ting2005taxonomy}. By plotting instantaneous current against voltage of 126 electric loads, they showed that the resulting Lissajous curves can be used for classifying load signatures. However, recent studies show that the image-based and V-I trajectory feature extraction techniques suffer from issues such as amplitude signal loss, and imperfect spatial structure features, resulting in higher training costs \cite{du2023nilm}. Some works are based on the combination of features such as steady-state and transient electrical characteristics. For example, Bao et al. \cite{bao2021feature} started with 20 extracted features (14 steady-state and 6 transient) and proposed a method to filter and rank them. After redundancy removal using mutual information and CRITIC weighting, and relevancy ranking using the Relief-F algorithm, they selected a final optimal subset of six features from a low-frequency dataset, including the third harmonic principal component, current crest factor, transient duration, mean active and reactive power, and equivalent impedance. 

The release of high-frequency datasets—notably PLAID \cite{gao2014plaid}, WHITED \cite{kahl2016whited}, and BLUED \cite{filip2011blued} supported the development of advanced features like FFT-based magnitude and phase harmonics. For instance, EMI features can help distinguish appliances of the same model, while harmonic features are effective for separating appliances with similar power consumption. In \cite{patri2014extracting}, a shapelet-based feature extraction approach on the BLUED dataset achieved an event detection accuracy of 98\%. Studies such as \cite{tabanelli2020feature} combined high-frequency features, such as odd current harmonics (up to the 10th order), with low-frequency features to enable NILM implementation on the edge devices. This compact feature set achieved up to 95.99\% classification accuracy while reducing memory usage to 9.4kB and computational load to 16K multiply–accumulate operations. Kahl et al. \cite{kahl2017comprehensive} conducted an extensive benchmarking study on 36 spectral and temporal features extracted from high-frequency datasets. Their analysis identified the most effective feature combinations for appliance recognition. However, they also showed that even the best-performing features are not always robust to realistic conditions, such as concurrent appliance operation, measurement noise, and different brands of appliances. In such cases, a promising alternative is using machine learning and deep learning models which can automatically extract discriminative representations from raw or preprocessed current-voltage waveforms. In \cite{kahl2022representation}, the authors compared three deep neural networks—an autoencoder, a convolutional autoencoder, and an end-to-end convolutional neural network (CNN)—trained on raw data, with a classical machine learning model that used 212 handcrafted features. The CNN obtained the $F1$ score of 0.75 on UK-DALE and 0.86 on BLOND, indicating that automatic feature extraction in deep learning models can perform better than conventional feature engineering methods. In the context of high-frequency NILM, authors in \cite{held2018frequency} propose using frequency invariant transformation of periodic signals (FIT-PS) as a feature which contains all information for a feedforward neural network, which leads to more robust results in scenarios with up to six active appliances at the same time. However, such networks with complicated architectures which are well performing are often heavy for deployment on edge devices. In practice, the model complexity must be reduced without compromising accuracy, which requires making the input data as compact and informative as possible. One effective way to achieve this is through dimensionality reduction techniques such as PCA and ICA. While PCA and ICA have been used in NILM only for data compression \cite{isanbaev2022comparative}, in this work we employ them as a feature extraction method. Specifically, we propose a fusion of PCA and ICA features combined with a ResNet-based classifier. Our results show that the proposed model performs better on high-frequency data than the state-of-the-art NILM architectures while still being lightweight for on-edge deployment.
 
\section{Datasets}\label{sec:dataset}
As we are focusing on signals with a high sampling rate to track the electrical properties of an appliance, we selected the PLAID dataset \cite{gao2014plaid}, which is measured at 30 kHz, with a total number of classes equal to 15. Each sample is a pair of voltage and current signals related to one particular class of appliances.

Since one of the goals of this research is to investigate the model’s performance on different number of simultaneously working appliances, we generated samples of aggregated signals where from one to $n_{max}$ (here $n_{max}$ = 15) types of appliances appear at the same time. For this, we developed a mixing algorithm that combines individual signals (sub-meter data) to create synthetic aggregated signals. This method implements Kirchhoff’s current law, ensuring that each synthetic signal is a physically valid superposition of individual appliance currents. The generation method is presented in Algorithm \ref{alg:composer}. This procedure is valid from a physical point of view as long as all the measurements are from the same power grid, i.e., the same voltage level, and frequency. 

Moreover, we accounted for the fact that some appliances may be duplicated from one to ten times. This is because one particular type of appliance in a household may be presented multiple times, e.g., three phone chargers, two air-conditioners, ten light bulbs. Aggregated samples were generated randomly, where each class has an equal probability of being presented in the mixture. By doing so, we aimed to reduce the class imbalance of the resulting dataset.

To prepare dataset, first, we resampled signals of the PLAID dataset to 3 kHz. Then we extracted around 19,000 regions of interest from the PLAID dataset. We split them into subsets in the following proportions 70\%, 10\% and 20\%, for train, validation and test sets respectively. Each aggregated sample is represented by a binary vector, where a value of 1 indicates that the corresponding appliance class is active, and included in the mixture. 

\begin{algorithm}
\caption{Random mixture method for synthetic aggregated data generation}
\label{alg:composer}
\KwIn{Labeled dataset $(X_{\mathrm{raw}}, y)$,  
combination range $n_{c} \in [n_{\min}, n_{\max}]$, 
minimum and maximum number of signatures $(f_{\min}, f_{\max})$.}
\KwOut{Aggregated dataset $(X, Y)$.}

\BlankLine
\begin{enumerate}[label=\protect\circled{\arabic*}]

\item\textbf{Initialization}\\
\Indp
\begin{itemize}
    \item Extract unique appliance classes $\mathcal{C} = \mathrm{unique}(y)$;\\
    \item For each class $c \in \mathcal{C}$, store sample indices:\\
    $\mathcal{D}_c = \{ i | y_i = c \}$;\\ 
\end{itemize}
\Indm

\BlankLine
\item\textbf{Generate class combinations}\\
\Indp
\begin{itemize}
    \item Randomly select unique combinations of size $n_{c}$:\\
    $\mathcal{Y}_{\text{comb}} = \{\, y_j \subseteq C \mid |y_j| = n_c \,\}$;\\
\end{itemize}

\Indm

\BlankLine
\item \textbf{For each combination $y_{j} \in \mathcal{Y}_{\text{comb}} $: }
\begin{itemize}
    \item Randomly assign vector $\mathbf{f} = [f_1, \ldots, f_{n_c}]$, where $f_i \sim \mathcal{U}(f_{\min}, f_{\max})$;
    \item Randomly sample $f_i$ indices from $\mathcal{D}_{y_i}$;  
    \item Form index set $I = \cup_i \text{samples}(\mathcal{D}_{y_i}, f_i)$; \\
    \item Compute aggregated signal $x = \sum_{k \in I} X_{\mathrm{raw}}[k]$;
    \item Collect labels $\mathbf{y} = \mathrm{unique}(\mathbf{y[I]})$;
    \item Append $(x, \mathbf{y})$ to $(X, Y)$.
\end{itemize}

\Return aggregated dataset $(X, Y)$.
\BlankLine

\end{enumerate}

\end{algorithm}

\section{Feature extraction}\label{sec:features}
This stage aims to extract the most informative features to identify the operational states of appliances, which is on/off states in the case of linear loads and also transition events for non-linear loads. PCA and ICA are dimensionality reduction techniques that also have been used for feature extraction in other domains \cite{yu2008integration}. In NILM, this approach has only recently been explored in our previous work, where we proposed using features extracted by ICA.  

Their application to NILM has only recently been investigated. In our previous work, we proposed classifying appliances using ICA-extracted features, showing that they could capture statistically independent load characteristics \cite{Sahar}. Based on that, and inspired by the work of \cite{reza2016ica}, we first discuss PCA and ICA, and then use the fusion of both, hereafter referred to as NILM-ICPC, to extract the feature set.
 
\subsection{Principal Component Analysis (PCA)}
\label{PCA}
PCA is a statistical technique used to transform a set of correlated variables into a smaller set of linearly uncorrelated components while retaining most of the variance present in the original data \cite{pearson1901liii}. 
Given a dataset $X_{\mathrm{{raw}}} = [x_1, x_2, \ldots, x_m]^T \in \mathbb{R}^{m \times n}$, where $m$ denotes the number of samples and $n$ the number of features, the method proceeds as follows. First, the data are mean-centered:
\begin{equation}
\tilde{X} = X_{\mathrm{raw}} - \mu, \quad \text{where } \mu = \frac{1}{m}\sum_{i=1}^{m} x_i
\end{equation}

Then, the sample covariance matrix is computed as:
\begin{equation}
S = \frac{1}{m-1} \tilde{X}^T \tilde{X}
\end{equation}

The eigenvalue decomposition of $S$ yields:
\begin{equation}
S v_i = \lambda_i v_i, \quad i = 1, 2, \ldots, n
\end{equation}
where $v_i$ and $\lambda_i$ denote the $i$-th eigenvector and its corresponding eigenvalue, respectively. The eigenvectors form an orthonormal basis, satisfying $v_i^{T} v_j = \delta_{ij}$. Each eigenvalue $\lambda_i$ represents the variance explained by its corresponding component. The eigenvectors corresponding to the $k_{\mathrm{pca}}$ largest eigenvalues form the projection matrix:

\begin{equation}
W_{\mathrm{pca}} = [v_1, v_2, \ldots, v_{k{\mathrm{pca}}}]
\end{equation}
The data are then projected into the reduced feature space as:
\begin{equation}
X_{pca} = \tilde{X} W_{\mathrm{pca}}
\end{equation}

Here, $X_{pca} \in \mathbb{R}^{m \times k_{pca}}$ represents the transformed data retaining the directions of maximal variance. 
It must be mentioned that the resulting principal components are only \textit{uncorrelated}---not necessarily \textit{statistically independent}---which may limit their discriminative power when non-Gaussian or higher-order dependencies exist among signals. This limitation led us to further explore ICA and fusion of ICA and PCA features.

\subsection{Independent Component Analysis (ICA)}
ICA is a statistical and computational method for recovering source signals from observed linear mixtures. It has found widespread application in feature extraction, dimensionality reduction, blind source separation, speech enhancement, and face recognition systems \cite{ica}. ICA models the data as a linear combination of latent variables where both the sources and the mixing process are unknown. The latent variables are assumed to be statistically independent and non-Gaussian, though the non-Gaussianity assumption can be relaxed when temporal dependencies are present, as they provide additional information for separation \cite{pearlmutter1996maximum}. ICA is well-suited to the NILM problem since the aggregated current or voltage signal can be approximated as a linear mixture of several independent sources (appliances) which can be expressed as:

\begin{equation}
x(t) = \sum_{k=1}^{K} a_k s_k(t) + \epsilon(t),
\end{equation}

where $x(t)$ is the aggregated signal, $s_k(t)$ represents the individual source signal of the $k$-th appliance, $a_k$ denotes the amplitude of each signature for appliance $k$, and $\epsilon(t)$ accounts for measurement noise and errors. 

Given the mean-centered data matrix $\tilde{X} \in \mathbb{R}^{m \times n}$, ICA assumes the following linear model:
\begin{equation}
\tilde{X} = Z A^{T},
\end{equation}
where $Z = [z_1, z_2, \ldots, z_m]^T \in \mathbb{R}^{m \times n}$ contains the statistically independent latent sources, and $A \in \mathbb{R}^{n \times n}$ is an unknown mixing matrix. The objective of ICA is to estimate an unmixing matrix $U \approx A^{-1}$ such that:
\begin{equation}
\hat{Z} = \tilde{X} U,
\label{eq:unmixing}
\end{equation}
where $\hat{Z}$ approximates the original independent components.

Unlike PCA, ICA exploits higher-order statistics by maximizing measures of non-Gaussianity. A commonly used objective function is the negentropy. For each estimated independent component $\hat{z}_i$ (a column of $\hat{Z}$), negentropy is defined as:
\[
J(\hat{z}_i) = H(\hat{z}_{i,\text{gauss}}) - H(\hat{z}_i),
\]
where $H(\cdot)$ denotes differential entropy, $\hat{z}_{i,\text{gauss}}$ is a Gaussian random variable with the same mean and variance as $\hat{z}_i$, and $\hat{z}_{i} = \tilde{X}u_{i}$. 
Since Gaussian variables have maximum entropy for a given variance, negentropy is always non-negative and zero only for Gaussian distributions. 

The ICA problem can therefore be formulated as finding the unmixing matrix that maximizes the non-Gaussianity of the extracted components:
\begin{equation}
U^* = \arg\max_U \sum_{i=1}^n J(\tilde{X} u_i),
\end{equation}
where $u_i$ is the $i$-th column of $U$.

In this work, the \textit{FastICA} algorithm is employed due to its computational efficiency and stable convergence. FastICA operates on whitened data and iteratively updates each unmixing vector $u_i$ to maximize negentropy: 

       \begin{equation}
       u_i^{(t+1)} = \mathbb{E}\{\tilde{X}^{T} g(\tilde{X} u_i^{(t)})\} - \mathbb{E}\{g'(\tilde{X} u_i^{(t)})\} u_i^{(t)},
       \end{equation}
       followed by normalization:
       \begin{equation}
       u_i^{(t+1)} = \frac{u_i^{(t+1)}}{\|u_i^{(t+1)}\|},
       \end{equation}
   where $g(\cdot)$ is a nonlinear function. 

After convergence, the estimated independent components are obtained as in equation \ref{eq:unmixing}, where each column of $\hat{Z}$ corresponds to an independent component (an appliance signature). 

\subsection{Fusion of PCA and ICA Features (NILM-ICPC)}
In the context of NILM, PCA can separate the sources by decorrelating the signals, but it cannot perform as expected due to the fact that uncorrelated signals are not necessarily independent. On the other hand, applying only ICA to large datasets leads to slow model convergence \cite{berg2005real}. Thus, we assumed that the fusion of ICA and PCA provides a richer and more discriminative feature space compared to either PCA or ICA alone (see Section~\ref{sec:results}). This feature extraction approach is adapted from~\cite{reza2016ica}, and is referred to as NILM-ICPC hereafter. 

First, PCA is applied on the mean-centered input data, $\tilde{X}$, as explained in previous sections to obtain the PCA features, $X_{\mathrm{pca}}$. Next, ICA is performed independently on the same input data to extract $k_{\mathrm{ica}}$ statistically independent components, denoted as $\hat{Z} = [\hat{z}_1, \hat{z}_2, \ldots, \hat{z}_{k_{\mathrm{ica}}}]$.

Following the approach in \cite{reza2016ica}, the ICA components are ranked using their kurtosis values:
\begin{equation}
\kappa_j = \frac{\mathbb{E}\left[(\hat{z}_j - \mu_j)^4\right]}{\sigma_j^4} - 3, \quad j = 1, \ldots, k_{\mathrm{ica}},
\end{equation}
where $\mu_j$ and $\sigma_j$ are the mean and standard deviation of the $j$-th independent component $\hat{z}_j$. 

Components with the most negative kurtosis (sub-Gaussian distributions) are considered more discriminative and are therefore prioritized. Let $r \leq k_{\mathrm{ica}}$ selected components be:
\begin{equation}
X_{\mathrm{ica}} = [\hat{z}_{(1)}, \hat{z}_{(2)}, \ldots, \hat{z}_{(r)}],
\end{equation}
where $\kappa_{(1)} < \kappa_{(2)} < \ldots < \kappa_{(r)}$ denotes the sorted kurtosis values in ascending order.

The fused feature matrix, denoted as $X_{\mathrm{icpc}}$, is constructed by concatenating the retained PCA and selected ICA features:
\begin{equation}
X_{\mathrm{icpc}} = [X_{\mathrm{pca}},\, X_{\mathrm{ica}}] \in \mathbb{R}^{m \times (k_{\mathrm{{pca}} }+ r)}.
\end{equation}

Finally, the fused features are normalized to zero mean and unit variance:
\begin{equation}
\bar{X}_{\mathrm{icpc}} = \frac{X_{\mathrm{icpc}} - \mu_{\mathrm{icpc}}}{\sigma_{\mathrm{icpc}}},
\end{equation}
and used as input to the downstream deep residual network. For clarity, the proposed feature extraction process is also summarized in Algorithm~\ref{alg-ICPC}.

\begin{algorithm}[t]
\label{alg-ICPC}
\caption{NILM-ICPC Feature extraction method}
\label{alg:nilm_icpc}
\KwIn{Mean-centered data $\tilde{X} \in \mathbb{R}^{m \times n}$}
\KwOut{Normalized fused feature matrix $\bar{X}_{\mathrm{icpc}}$}

\begin{enumerate}[label=\protect\circled{\arabic*}]

\BlankLine

\item \textbf{Apply PCA} \\
\Indp $W_{\mathrm{pca}} \leftarrow \text{PCA}(\tilde{X})$ \\
$X_{\mathrm{pca}} \leftarrow \tilde{X} W_{\mathrm{pca}}$
\Indm

\item \textbf{Apply ICA} \\
\Indp $W_{\mathrm{ica}} \leftarrow \text{FastICA}(\tilde{X},\, k_{\mathrm{ica}})$ \\
$\hat{Z} \leftarrow \tilde{X} W_{\mathrm{ica}}$
\Indm

\item \textbf{Rank ICA components by kurtosis} \\
\Indp $\kappa_j \leftarrow \text{Kurtosis}(\hat{Z}[:,j])$ for $j = 1,\ldots,k_{\mathrm{ica}}$ \\
$\text{order} \leftarrow \text{Argsort}(\kappa)$ 
$\triangleright$ most negative first \\
$X_{\mathrm{ica}} \leftarrow \hat{Z}[:,\, \text{order}[1:r]]$
\Indm

\item \textbf{Fuse PCA and ICA features} \\
\Indp $X_{\mathrm{icpc}} \leftarrow [\, X_{\mathrm{pca}},\, X_{\mathrm{ica}} \,]$
\Indm

\item \textbf{Normalize fused features} \\
\Indp $\bar{X}_{\mathrm{icpc}} \leftarrow \dfrac{X_{\mathrm{icpc}} - \mu_{\mathrm{icpc}}}{\sigma_{\mathrm{icpc}}}$
\Indm

\Return $\bar{X}_{\mathrm{icpc}}$

\end{enumerate}
\end{algorithm}

\section{Algorithms and Models}\label{sec:algorithms}
\subsection{Proposed Model: Fusion-ResNet}

\textit{Fusion-ResNet} is based on our previous model presented in \cite{Sahar} while we introduce a new feature extraction method. The proposed model, extends the residual feed-forward architecture \cite{Sahar} by integrating feature fusion from PCA and ICA. The complete pipeline consists of four main stages: (1) feature extraction via PCA and ICA, (2) feature normalization and fusion, (3) deep residual feed-forward transformation, and (4) multi-label classification.

\begin{figure*}
\begin{center}
\centering
\centerline{\includegraphics[width=0.9\textwidth]{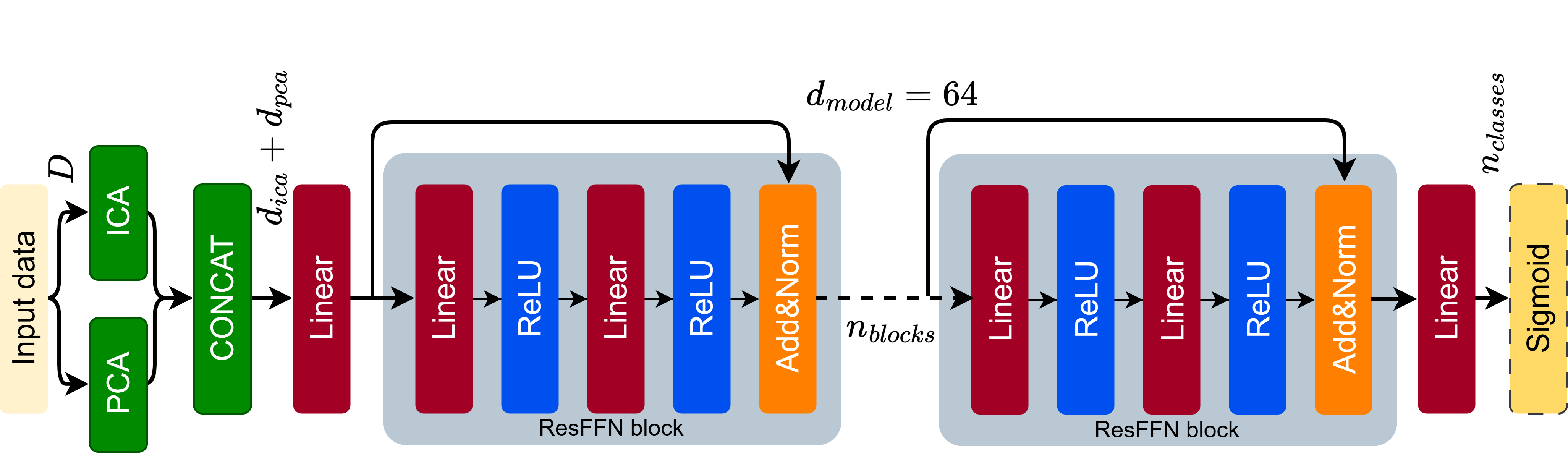}}
\caption{Architecture of the proposed model, Fusion-ResNet. The number of parameters for this model is 65,000.}
\label{proposed_model}
\end{center}
\end{figure*}

In fact, the fused feature matrix, $\bar{X}_{\mathrm{ICPC}}$ is fed into a residual feed-forward network composed of $n_{\text{blocks}} = 18$ residual blocks. Each \textbf{ResFFN} block contains two fully connected layers with ReLU activations and dropout ($p = 0.1$), followed by residual addition and layer normalization as illustrated in Fig.~\ref{proposed_model}. The final linear layer projects the latent representation $h_L$ to a vector of appliance activation logits
\begin{equation}
\hat{y} = W_o h_L + b_o \in \mathbb{R}^{n_{\text{c}}},
\end{equation}
where each element $\hat{y}_{j}$ corresponds to the predicted likelihood of appliance $j$ being active.

\paragraph {Training objective} 
The model is trained using the Binary Cross-Entropy (BCE) loss:
\begin{eqnarray}
\nonumber
\mathcal{L}_{\text{BCE}} =
&-&\frac{1}{N n_{c}} \sum_{i=1}^{N}\sum_{j=1}^{n_c}
\left[y_{ij}\log\sigma(\hat{y}_{ij})\right.\\ 
 &+& \left. (1 - y_{ij})\log(1-\sigma(\hat{y}_{ij}))\right],
\end{eqnarray}
where $y_{ij}\in\{0,1\}$ denotes the ground-truth activation of appliance $j$ in sample $i$, $N$ is the number of samples, and $n_{c}$ is the number of classes. 

\paragraph{Inference} During inference, each output $\hat{y}_{ij}$ is passed through the sigmoid activation function
\begin{equation}
   \sigma(\hat{y}_{ij}) = \frac{1}{1 + e^{-\hat{y}_{ij}}}, 
\end{equation}

which maps the predicted logit values to probabilities in $(0, 1)$. 
An appliance is classified as \emph{active} if $\sigma(\hat{y}_{ij}) > \tau$, 
where $\tau = 0.5$ is the fixed threshold used throughout all experiments. 

\paragraph{Optimization and training}
The model is optimized using the Adam optimizer (learning rate $10^{-2}$). Training runs for up to $150$ epochs with a batch size of $128$. 

\paragraph{Evaluation}
Model performance is evaluated using the sample-averaged $F_1$ score, which involves true and false positives (TP and FP respectively) as well as true and false negatives (TN and FN respectively). The $F_1$ score for $i$-th sample is given by: 

\begin{equation}
    F_{1,i} = \frac{2\,\mathrm{TP}_{i}}{2\,\mathrm{TP}_{i} + \mathrm{FP}_{i} + \mathrm{FN}_{i}}.
    \label{eq:f1score}
\end{equation}

The final score is obtained by averaging $F_{1,i}$ across all $N$ test samples:
\begin{equation}
    F_1 = \frac{1}{N}\sum_{i=1}^{N} F_{1,i},
\end{equation}

\subsection{Baseline models}
\subsubsection{FIT-PS+LSTM}
The FIT-PS method is a novel signal processing method, which was successfully applied as a feature extraction method for classification in NILM \cite{held2018frequency}. It consists of three steps. The first step is to divide the sampled signal with respect to the fundamental frequency of the power grid. The division is done by finding the abscissa crossing, namely, the change from negative values of voltage to positive values. Abscissa crossing is chosen since it represents the point of maximum steepness, has an almost constant derivative in sinusoidal signals, and is less affected by amplitude variations compared to other parts of the signal. In the second step, abscissa crossing is used to estimate the linear approximation of absolute position of zero crossing. Next, linear interpolation is applied to assign indices to each individual period of the signal. The resulting matrix $X_{l,k}$ has $n_l \times n_k$ dimensions, where $n_l$ is the number of periods and $n_k$ is the number of sampling points in each period.

The signal passed through the FIT-PS method can be fed to the LTSM network, where $n_k$ is the input dimension.

An output of such a network is being averaged across a number of periods and then being passed through a fully connected layer with output size equal to $n_{\mathrm{classes}}$. Finally, a sigmoid layer is used to calculate scores of each class being present in the aggregated signal.  

\subsubsection{Fryze+CNN}
The CNN model proposed in \cite{faustine2020multi} uses the Fryze power theory \cite{staudt2008fryze} and the Euclidean distance matrix as feature extraction step for the multi-label classifier. Within the theory, the activation current is decomposed into orthogonal components related to electrical energy in the time-domain:
\begin{equation}
i(t)=i(t)_a+i(t)_f
\end{equation}
The active current $i(t)_{a}$ is the current passing through the resistive load. In Fryze’s theory, the active power is calculated as the average value of $i(t) \cdot v(t)$ over one fundamental cycle $T_{s}$ defined as follows;

\begin{equation}
i(t)_a=\frac{p_a}{v_{r m s}^2} v(t)
\end{equation}

where the rms voltage $v_{rms}$ is expressed as follows
\begin{equation}
v_{r m s}=\sqrt{\frac{1}{T_s} \sum_{t=1}^{T_s} v(t)^2}
\end{equation}

The non-active component is then equal to
\begin{equation}
i(t)_f=i(t)-i(t)_a
\end{equation}
The orthogonal components of the current, namely $i(t)_a$, and $i(t)_f$ undergo two pre-processing steps. Firstly, the signals are dimensionally reduced using the piece-wise aggregate approximation. Secondly, the distance matrix for each signal is computed. The two resulting distance matrices are then combined to create input for the multi-label classifier. The classifier is a four-block convolutional neural network, featuring 16, 32, 64 and 128 channels, with kernel sizes of $5 \times 5$, $5 \times 5$, $3 \times 3$, and $3 \times 3$ and strides of size 2, respectively. The remaining part of the network comprises three linear layers, consisting of 512, 1024 and $n_{\mathrm{classes}}$ neurons, respectively. The main activation function employed is ReLU.

\section{Experiments}\label{sec:experiments}
For the case study, we trained and validated the following models: 
ResNetFFN with ICA features, with PCA features and then with NILM-ICPC features, as well as Fryze+CNN \cite{faustine2020multi}, and FIT-PS+LSTM \cite{held2018frequency}. The experiment required 20 minutes of processing time on a machine with $2 \times$ RTX 2080 Ti GPUs and 128GB of RAM. 

We assessed the performance of classification algorithms using the $F_1$ score, averaged over samples. To assess the performance of models under stress conditions, we computed the $F_1$ score for varying numbers of appliances working simultaneously, from 1 to $n_{\mathrm{classes}}$. The distribution of $F_1$ score across different number of simultaneously working appliances is uniform. This implies that there is no model bias towards a specific number of appliances during a runtime. Intuitively, as the number of appliances grows, the $F_1$ score is expected to decrease. This is mainly because low-power appliances, such as chargers, are regarded as noise when they operate simultaneously with high-power appliances. 

\section{Results}\label{sec:results}
\begin{figure}[h]
    \centering
    \includegraphics[width=0.9\columnwidth]{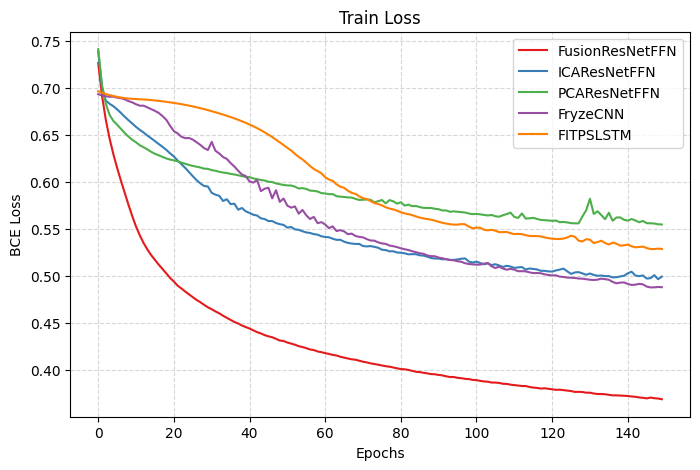}
    \caption{BCE loss for the training set.}
\label{fig:loss_train}
\end{figure}

\begin{figure}[h]
    \centering
    \includegraphics[width=0.9\columnwidth]{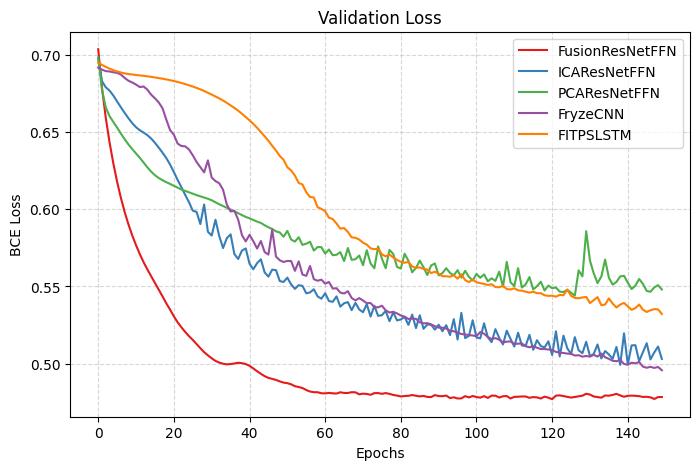}
    \caption{BCE loss for the validation set}
\label{fig:loss_val}
\end{figure}

\begin{figure}[h]
    \centering
    \includegraphics[width=0.9\columnwidth]{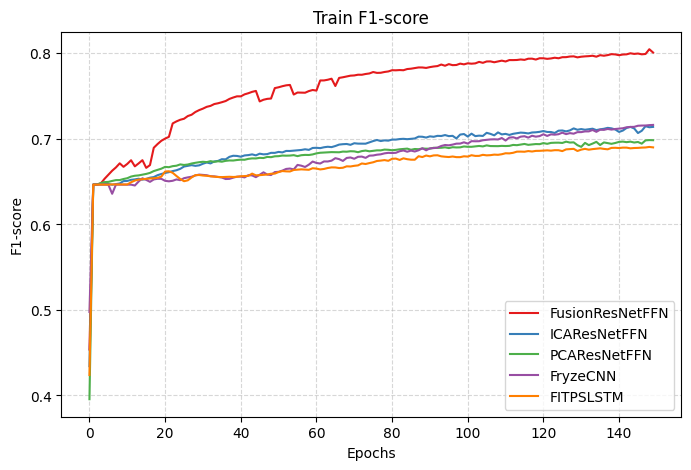}
    \caption{$F_1$ score (sample average) for train set.}
\label{fig:f1-train}
\end{figure}

\begin{figure}[h]
    \centering
    \includegraphics[width=0.9\columnwidth]{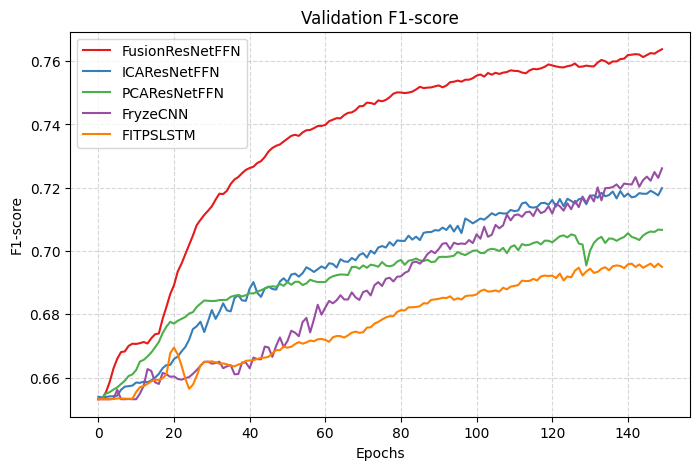}
    \caption{$F_1$ score (sample average) for validation set.}
\label{fig:f1-val}
\end{figure}

Figures \ref{fig:loss_train} and \ref{fig:loss_val} show the binary cross-entropy loss for training and validation of the aforementioned models over 150 epochs. 
The Fusion-ResNet model with feed forward network (Fusion-ResNetFFN) with NILM-ICPC features achieves the lowest training and validation losses. The $F_1$ scores of the training and validation sets depicted in Figs.~\ref{fig:f1-train} and \ref{fig:f1-val} respectively, clearly show that Fusion-ResNetFFN again outperforms other architectures. 

The results are summarized in Table \ref{tab:my-table} where the $F_1$ scores (sample averaging) of all models are compared. Fusion-ResNetFFN achieves the highest average $F_1$ score of 0.77 with relatively low training time (73.5 s) and inference cost (0.0017 ms/sample). These results indicate that the proposed NILM-ICPC feature set enhances the model performance compared to using ICA- or PCA-based features alone. Furthermore, other deep models such as Fryze+CNN and FIT-PS+LSTM exhibit significantly longer training and inference times due to their higher architectural complexity while having lower accuracy.

\begin{table}
\centering
\caption{Average $F1$ score for all models}
\label{tab:my-table}
\begin{tabular}{lcccccc}
\toprule
\textbf{Model} &
  \textit{F1 score} &
  \textit{Train time (s)} &
\textit{Inference (ms/sample)} \\
\midrule
\textit{Fusion-ResNetFFN} & 0.77 & 73.5 & 0.0017\\
\textit{ICA-ResNetFFN} & 0.73 & 73.5 & 0.0017\\
\textit{PCA-ResNetFFN} & 0.72 & 72.9  & 0.0017\\
\textit{Fryze+CNN} & 0.74 & 615.2 & 0.26\\
\textit{FIT-PS+LSTM} & 0.71 & 92.6 & 0.004\\

\bottomrule
\end{tabular}
\end{table}

\begin{figure*}[t]
\centering
\includegraphics[width=\textwidth]{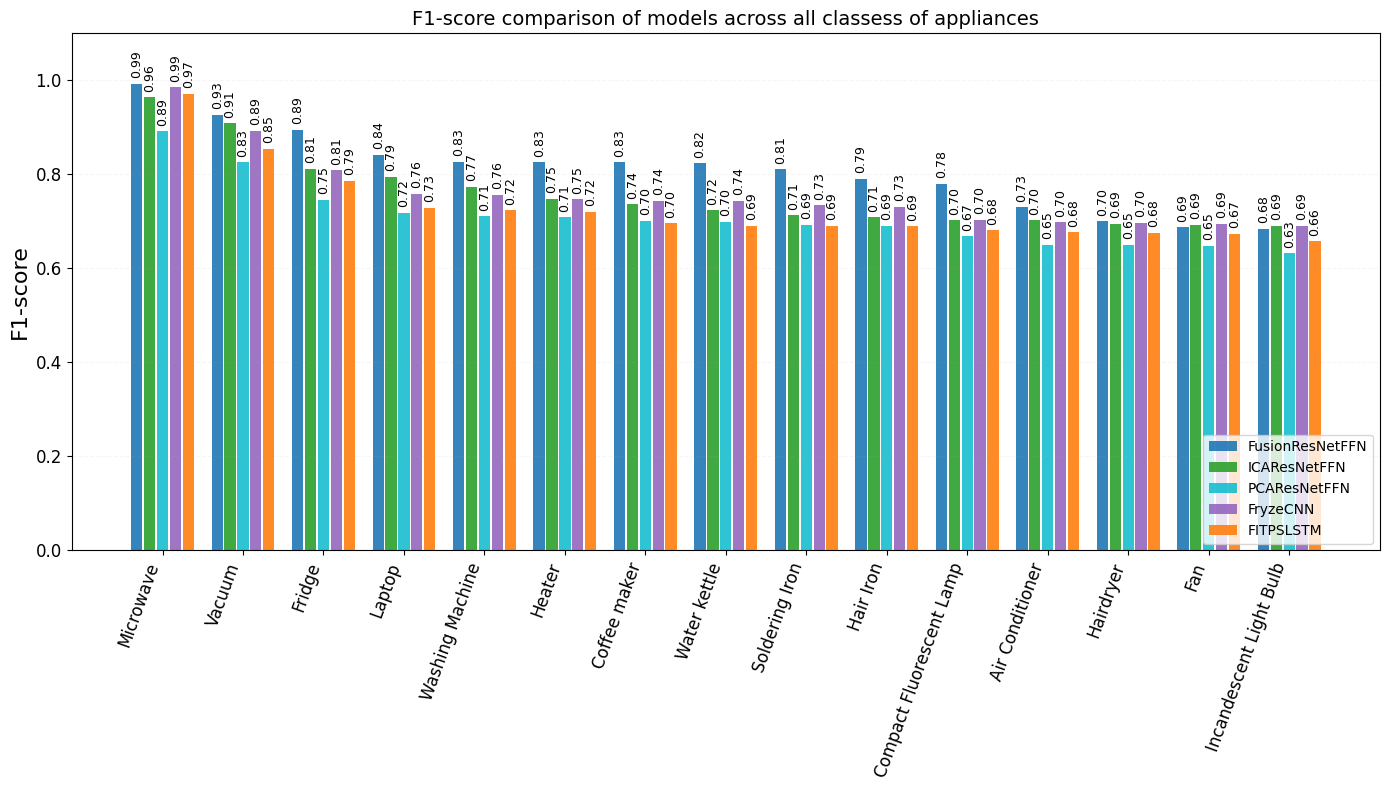}
\caption{Distribution of $F_1$ score (samples averaging) across different classes of appliances.}
\label{fig:per_class}
\end{figure*}

The $F1$ scores of all models across all classes of appliances (15 classes) are depicted in Fig.~\ref{fig:per_class}. All models can detect high-power appliances such as the microwave and vacuum almost perfectly, reaching $F1$ score above 90\%, while for other devices such as fans, hairdryers, and fluorescent lamps, all models perform below 70\%. It most be noted that the Fusion-ResNetFFN achieves the highest overall $F1$ score compared to other models. These results show the model’s robustness across different load types. However, no single architecture performs best for all appliances. There is a clear trade-off between accuracy and generalization. A practical extension would be to combine the best-performing models for each appliance type into an ensemble, though this would increase computational complexity.

\begin{figure*}
\centering
\includegraphics[width=\textwidth]{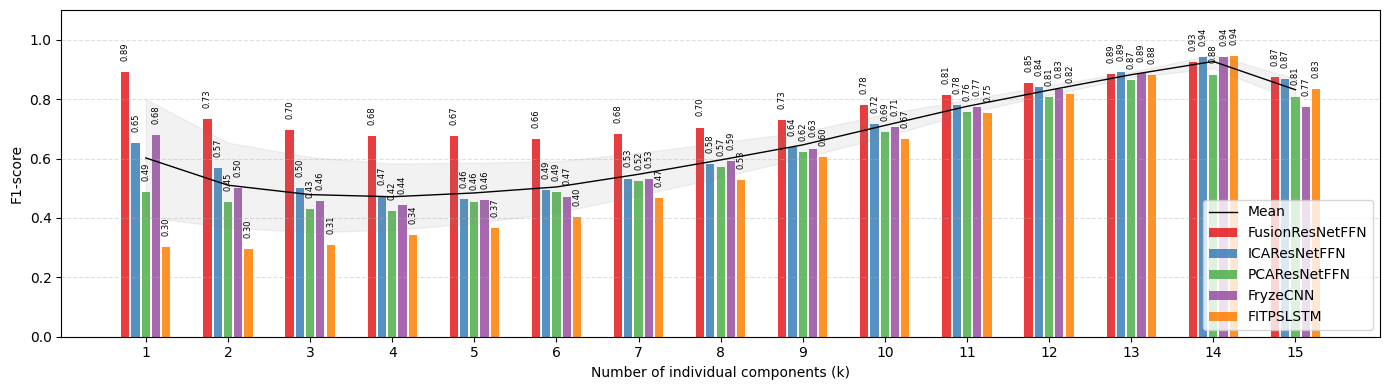}
\caption{Distribution of $F_1$ score (samples averaging) across different number of simultaneously working appliances.}
\label{fig:simultan-f1-bars}
\end{figure*}

Further, we evaluate how the model performance changes as the number of simultaneously operating appliances increases. For this, we increase the number of active appliances from 1 to 15, where all selected appliances operate concurrently, as illustrated in Fig.~\ref{fig:simultan-f1-bars}. Fusion resnet outperformed all other baseline models across varying number of active appliances which shows that our proposed model is even robust to stress condition compared to others. 

\begin{figure}
\centering
\includegraphics[width=\columnwidth]{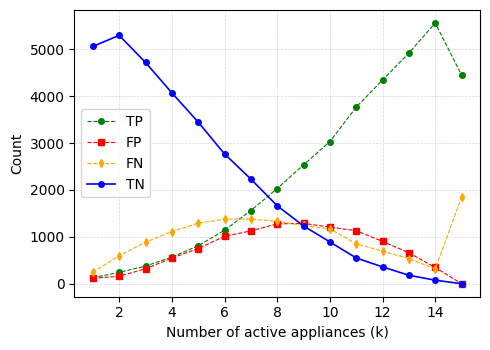}
\caption{Confusion matrix components for varying number of active appliances $k$.}
\label{fig:confusion}
\end{figure}

We also notice that as the number of simultaneously operating appliances $k$ increases, the total number of possible combinations of active appliances increases combinatorially: $\sum_{k=1}^{n} \binom{n}{k}$ with $n$ being the total number of available appliances. On average (Fig. \ref{fig:simultan-f1-bars}, black line), the accuracy of the models decreases as the number of active appliances increases up to approximately $k= 7\text{–}8$, i.e., when about half of the appliances are running. This is due to the overlap in feature space that grows proportionally to the combinatorial probabilities (e.g, for $k = 7$, there are $6435$ combinations of active appliances). For larger values ($ 8 \leq k \leq 14$), the $F1$ score increases again. However, this does not necessarily indicate higher model performance; rather, it is related to the metric behavior. For this, we plotted Fig.~\ref{fig:confusion}, which presents the counts of the confusion matrix components, and shows that as more devices are active, the number of true positives (TPs) naturally increases, while true negatives (TNs) decrease, leading to higher values of $F1$ score (See Eq.~(\ref{eq:f1score})). 

Interestingly, we see that for $k = 15$, every appliance is active, implying there is only one combination possible (all appliances ``on'') and hence no diversity ($N_{15} = 1$) with all samples having identical labels ($Y_{i}= 1$). Therefore, $k=15$ can be considered as the extreme case where both the number of false negatives (FN) and true positives (TP) drop sharply due to the lack of diversity. As a consequence, it should be taken into account that the $F_1$ score depends on the diversity of appliance combinations for a specific value of $k$. 

\section{Conclusion}\label{sec:conclusion}
We presented an end-to-end framework for multi-label high-frequency NILM classification that integrates a novel feature extraction method and a lightweight neural architecture. Our results show the importance of considering the underlying physics of the data when selecting an appropriate model to accurately capture the physical characteristics of the energy disaggregation problem. In this regard, we employed fusion of the features extracted by ICA and PCA. A primary objective of this research was to evaluate the performance of the algorithm in scenarios with varying numbers of concurrent appliances. To ensure that our dataset was sufficiently diverse to include a large set of combinations, we deliberately curated a rich and comprehensive dataset. Our findings indicate that our model using the proposed feature set outperforms existing baseline models, and that it is computationally more efficient. Given its compact architecture (65k parameters), the model shows potential for being deployed on on-edge devices. Therefore, future work will focus on integrating the proposed framework with embedded hardware platforms and evaluating its real-time performance.   

\bibliography{references}

\begin{thebibliography}{10}
\providecommand{\url}[1]{#1}
\csname url@samestyle\endcsname
\providecommand{\newblock}{\relax}
\providecommand{\bibinfo}[2]{#2}
\providecommand{\BIBentrySTDinterwordspacing}{\spaceskip=0pt\relax}
\providecommand{\BIBentryALTinterwordstretchfactor}{4}
\providecommand{\BIBentryALTinterwordspacing}{\spaceskip=\fontdimen2\font plus
\BIBentryALTinterwordstretchfactor\fontdimen3\font minus
  \fontdimen4\font\relax}
\providecommand{\BIBforeignlanguage}[2]{{%
\expandafter\ifx\csname l@#1\endcsname\relax
\typeout{** WARNING: IEEEtran.bst: No hyphenation pattern has been}%
\typeout{** loaded for the language `#1'. Using the pattern for}%
\typeout{** the default language instead.}%
\else
\language=\csname l@#1\endcsname
\fi
#2}}
\providecommand{\BIBdecl}{\relax}
\BIBdecl

\bibitem{angelis2022nilm}
G.-F. Angelis, C.~Timplalexis, S.~Krinidis, D.~Ioannidis, and D.~Tzovaras,
  ``Nilm applications: Literature review of learning approaches, recent
  developments and challenges,'' \emph{Energy and Buildings}, vol. 261, p.
  111951, 2022.

\bibitem{Event_detection}
K.~D. Anderson, M.~E. Bergés, A.~Ocneanu, D.~Benitez, and J.~M. Moura, ``Event
  detection for non intrusive load monitoring,'' in \emph{IECON 2012 - 38th
  Annual Conference on IEEE Industrial Electronics Society}, 2012, pp.
  3312--3317.

\bibitem{hart1992nonintrusive}
G.~W. Hart, ``Nonintrusive appliance load monitoring,'' \emph{Proceedings of
  the IEEE}, vol.~80, no.~12, pp. 1870--1891, 1992.

\bibitem{en13092195}
H.~Rafiq, X.~Shi, H.~Zhang, H.~Li, and M.~K. Ochani, ``A deep recurrent neural
  network for non-intrusive load monitoring based on multi-feature input space
  and post-processing,'' \emph{Energies}, vol.~13, no.~9, 2020.

\bibitem{chang2015feature}
H.-H. Chang, M.-C. Lee, N.~Chen, C.-L. Chien, and W.-J. Lee, ``Feature
  extraction based hellinger distance algorithm for non-intrusive aging load
  identification in residential buildings,'' in \emph{2015 IEEE Industry
  Applications Society Annual Meeting}.\hskip 1em plus 0.5em minus 0.4em\relax
  IEEE, 2015, pp. 1--8.

\bibitem{kamyshev2025cold}
I.~Kamyshev, S.~M. Hoosh, D.~Kriukov, E.~Gryazina, and H.~Ouerdane, ``Cold:
  Concurrent loads disaggregator for non-intrusive load monitoring,'' in
  \emph{2025 IEEE Kiel PowerTech}.\hskip 1em plus 0.5em minus 0.4em\relax IEEE,
  2025, pp. 1--6.

\bibitem{Rafiq2021GeneralizabilityIO}
H.~Rafiq, X.~Shi, H.~Zhang, H.~Li, M.~K. Ochani, and A.~A. Shah,
  ``Generalizability improvement of deep learning-based non-intrusive load
  monitoring system using data augmentation,'' \emph{IEEE Transactions on Smart
  Grid}, vol.~12, pp. 3265--3277, 2021.

\bibitem{Sahar}
S.~M. Hoosh, I.~Kamyshev, and H.~Ouerdane, ``Enhancing non-intrusive load
  monitoring with features extracted by independent component analysis,'' in
  \emph{2025 7th International Youth Conference on Radio Electronics,
  Electrical and Power Engineering (REEPE)}, 2025, pp. 01--06.

\bibitem{sultanem2002using}
F.~Sultanem, ``Using appliance signatures for monitoring residential loads at
  meter panel level,'' \emph{IEEE Transactions on Power Delivery}, vol.~6,
  no.~4, pp. 1380--1385, 2002.

\bibitem{leeb1995transient}
S.~B. Leeb, S.~R. Shaw, and J.~L. Kirtley, ``Transient event detection in
  spectral envelope estimates for nonintrusive load monitoring,'' \emph{IEEE
  Transactions on Power Delivery}, vol.~10, no.~3, pp. 1200--1210, 1995.

\bibitem{ting2005taxonomy}
K.~Ting, M.~Lucente, G.~S. Fung, W.~Lee, and S.~Hui, ``A taxonomy of load
  signatures for single-phase electric appliances,'' in \emph{IEEE PESC (Power
  Electronics Specialist Conference)}, 2005, pp. 12--18.

\bibitem{du2023nilm}
Z.~Du, B.~Yin, Y.~Zhu, X.~Huang, and J.~Xu, ``A nilm load identification method
  based on structured vi mapping,'' \emph{Scientific Reports}, vol.~13, no.~1,
  p. 21276, 2023.

\bibitem{bao2021feature}
S.~Bao, L.~Zhang, X.~Han, W.~Li, D.~Sun, Y.~Ren, N.~Liu, M.~Yang, and B.~Zhang,
  ``Feature selection method for nonintrusive load monitoring with balanced
  redundancy and relevancy,'' \emph{IEEE Transactions on Industry
  Applications}, vol.~58, no.~1, pp. 163--172, 2021.

\bibitem{gao2014plaid}
J.~Gao, S.~Giri, E.~C. Kara, and M.~Berg{\'e}s, ``Plaid: a public dataset of
  high-resoultion electrical appliance measurements for load identification
  research: demo abstract,'' in \emph{proceedings of the 1st ACM Conference on
  Embedded Systems for Energy-Efficient Buildings}, 2014, pp. 198--199.

\bibitem{kahl2016whited}
M.~Kahl, A.~U. Haq, T.~Kriechbaumer, and H.-A. Jacobsen, ``Whited-a worldwide
  household and industry transient energy data set,'' in \emph{3rd
  International Workshop on Non-Intrusive Load Monitoring}, 2016, pp. 1--4.

\bibitem{filip2011blued}
A.~Filip \emph{et~al.}, ``Blued: A fully labeled public dataset for event-based
  nonintrusive load monitoring research,'' in \emph{2nd workshop on data mining
  applications in sustainability (SustKDD)}, vol. 2012, 2011.

\bibitem{patri2014extracting}
O.~P. Patri, A.~V. Panangadan, C.~Chelmis, and V.~K. Prasanna, ``Extracting
  discriminative features for event-based electricity disaggregation,'' in
  \emph{2014 IEEE Conference on Technologies for Sustainability
  (SusTech)}.\hskip 1em plus 0.5em minus 0.4em\relax IEEE, 2014, pp. 232--238.

\bibitem{tabanelli2020feature}
E.~Tabanelli, D.~Brunelli, and L.~Benini, ``A feature reduction strategy for
  enabling lightweight non-intrusive load monitoring on edge devices,'' in
  \emph{2020 IEEE 29th International Symposium on Industrial Electronics
  (ISIE)}.\hskip 1em plus 0.5em minus 0.4em\relax IEEE, 2020, pp. 805--810.

\bibitem{kahl2017comprehensive}
M.~Kahl, A.~Ul~Haq, T.~Kriechbaumer, and H.-A. Jacobsen, ``A comprehensive
  feature study for appliance recognition on high frequency energy data,'' in
  \emph{Proceedings of the Eighth International Conference on Future Energy
  Systems}, 2017, pp. 121--131.

\bibitem{kahl2022representation}
M.~Kahl, D.~Jorde, and H.-A. Jacobsen, ``Representation learning for appliance
  recognition: A comparison to classical machine learning,'' \emph{arXiv
  preprint arXiv:2209.03759}, 2022.

\bibitem{held2018frequency}
P.~Held, S.~Mauch, A.~Saleh, D.~O. Abdeslam, and D.~Benyoucef, ``Frequency
  invariant transformation of periodic signals (fit-ps) for classification in
  nilm,'' \emph{IEEE Transactions on Smart Grid}, vol.~10, no.~5, pp.
  5556--5563, 2018.

\bibitem{isanbaev2022comparative}
V.~Isanbaev, R.~Ba{\~n}os, F.~M. Arrabal-Campos, C.~Gil, F.~G. Montoya, and
  A.~Alcayde, ``A comparative study on pretreatment methods and dimensionality
  reduction techniques for energy data disaggregation in home appliances,''
  \emph{Advanced Engineering Informatics}, vol.~54, p. 101805, 2022.

\bibitem{yu2008integration}
S.-N. Yu and K.-T. Chou, ``Integration of independent component analysis and
  neural networks for ecg beat classification,'' \emph{Expert systems with
  applications}, vol.~34, no.~4, pp. 2841--2846, 2008.

\bibitem{reza2016ica}
M.~S. Reza and J.~Ma, ``Ica and pca integrated feature extraction for
  classification,'' in \emph{2016 IEEE 13th international conference on signal
  processing (ICSP)}.\hskip 1em plus 0.5em minus 0.4em\relax IEEE, 2016, pp.
  1083--1088.

\bibitem{pearson1901liii}
K.~Pearson, ``Liii. on lines and planes of closest fit to systems of points in
  space,'' \emph{The London, Edinburgh, and Dublin philosophical magazine and
  journal of science}, vol.~2, no.~11, pp. 559--572, 1901.

\bibitem{ica}
C.~Jutten and J.~Herault, ``Blind separation of sources, part i: An adaptive
  algorithm based on neuromimetic architecture,'' \emph{Signal processing},
  vol.~24, no.~1, pp. 1--10, 1991.

\bibitem{pearlmutter1996maximum}
B.~Pearlmutter and L.~Parra, ``Maximum likelihood blind source separation: A
  context-sensitive generalization of ica,'' \emph{Advances in neural
  information processing systems}, vol.~9, 1996.

\bibitem{berg2005real}
M.~Berg and E.~Bondesson, ``Real-time implementation of a combined pca-ica
  algorithm for blind source separation,'' 2005.

\bibitem{faustine2020multi}
A.~Faustine and L.~Pereira, ``Multi-label learning for appliance recognition in
  nilm using fryze-current decomposition and convolutional neural network,''
  \emph{Energies}, vol.~13, no.~16, p. 4154, 2020.

\bibitem{staudt2008fryze}
V.~Staudt, ``Fryze-buchholz-depenbrock: A time-domain power theory,'' in
  \emph{2008 International School on Nonsinusoidal Currents and
  Compensation}.\hskip 1em plus 0.5em minus 0.4em\relax IEEE, 2008, pp. 1--12.

\end{thebibliography}
\bibliographystyle{IEEEtran}
\end{document}